\documentclass{ieeeaccess}
\usepackage{cite,graphicx,color,url}
\usepackage{algorithm,algpseudocode}
\usepackage{amssymb,amsmath,mathtools}
\usepackage{amsfonts}
\usepackage{caption}

\usepackage{bm,bbm}
\usepackage{booktabs}
\usepackage[caption=false,font=footnotesize]{subfig}
\graphicspath{{./_figs/}}

\def\BibTeX{{\rm B\kern-.05em{\sc i\kern-.025em b}\kern-.08em
    T\kern-.1667em\lower.7ex\hbox{E}\kern-.125emX}}
\begin{document}
\history{Date of publication xxxx 00, 0000, date of current version xxxx 00, 0000.}
\doi{10.1109/ACCESS.2017.DOI}

\title{Communication-oriented Model Fine-tuning for Packet-loss Resilient Distributed Inference under Highly Lossy IoT Networks}
\author{
	\uppercase{Sohei~Itahara}\authorrefmark{1}, \IEEEmembership{Student~Member,~IEEE},
	\uppercase{Takayuki~Nishio}\authorrefmark{2}, \IEEEmembership{Senior~Member,~IEEE},
	\uppercase{Yusuke~Koda}\authorrefmark{3}, \IEEEmembership{Graduate~Student~Member,~IEEE},
		and \uppercase{Koji~Yamamoto}.\authorrefmark{1}, \IEEEmembership{Senior~Member,~IEEE}
}
\address[1]{Graduate School of Informatics, Kyoto University, Kyoto 606-8501, Japan. (e-mail: kyamamot@i.kyoto-u.ac.jp)}
\address[2]{School of Engineering, Tokyo Institute of Technology Ookayama, Meguro-ku, Tokyo, 152-8550, Japan. (e-mail: nishio@ict.e.titech.ac.jp)}
\address[3]{Centre of Wireless Communications, University of Oulu, 90014 Oulu, Finland. (e-mail: Yusuke.Koda@oulu.fi)}
\tfootnote{This work was supported in part by JST PRESTO Grant Number JPMJPR2035.}

\markboth
{Author \headeretal: Preparation of Papers for IEEE TRANSACTIONS and JOURNALS}
{Author \headeretal: Preparation of Papers for IEEE TRANSACTIONS and JOURNALS}

\corresp{Corresponding author: Takayuki~Nishio (e-mail: nishio@ict.e.titech.ac.jp).}

\begin{abstract}
	The distributed inference (DI) framework has gained traction as a technique for real-time applications empowered by cutting-edge deep machine learning (ML) on resource-constrained Internet of things (IoT) devices.
	In DI, computational tasks are offloaded from the IoT device to the edge server via lossy IoT networks.
	However, generally, there is a communication system-level trade-off between communication latency and reliability;
	thus, to provide accurate DI results, a reliable and high-latency communication system is required to be adapted, which results in non-negligible end-to-end latency of the DI.
	This motivated us to improve the trade-off between the communication latency and accuracy by efforts on ML techniques.
	Specifically, we have proposed a communication-oriented model tuning (COMtune), which aims to achieve highly accurate DI with low-latency but unreliable communication links.
	In COMtune, the key idea is to fine-tune the ML model by emulating the effect of unreliable communication links through the application of the dropout technique.
	This enables the DI system to obtain robustness against unreliable communication links.
	Our ML experiments revealed that COMtune enables accurate predictions with low latency and under lossy networks.
\end{abstract}


\titlepgskip=-15pt

\maketitle

\section{Introduction}
\label{sec:introduction}
\PARstart{T}{he} Internet of things (IoT) is employed to enable multiple novel applications
 by combining the physical sensing of IoT devices with deep learning-based data analysis. 
Although deep learning technology is developing rapidly, it satisfying privacy and latency demands of the applications on resource-constrained IoT systems continue to pose a challenge. 
For example, factory automation and smart grids require latency of less than 10\,ms and 20\,ms, respectively~\cite{philipp2017latency}.
In contrast, in smart home applications, IoT sensors such as visual and audio sensors obtain privacy-sensitive data that should not be exposed~\cite{huichen2016iot}.

Distributed inference (DI) frameworks have been researched to address the privacy and latency challenges of deep learning deployment on IoT systems.
In the DI framework for deep neural networks (DNNs)~\cite{nicholas2015candeep,wang2018not}, computationally expensive tasks are offloaded from the IoT devices to the locally located edge servers to reduce computation latency and the risk of data leakage, as compared with cloud computing.
In the DI framework, the IoT devices and the edge server collaboratively process the portion of DNN, otherwise known as sub-DNN, by exchanging messages (e.g., outputs of sub-DNN) via IoT networks.
Details of the DI are explained as follows:
A well-trained DNN is divided into sub-DNNs through layers.
The IoT device stores the input-side sub-DNN, while the edge server stores the output-side sub-DNN.
The device obtains the output of the sub-DNN (i.e., the activations of the original DNN) from the raw input.
Next, the activation is transmitted to the edge server, and the server generates an inference result from the activations using its sub-DNN.

Although the DI reduces the computation latency and preserves the data privacy,
the problem of the communication latency of the DI is posed~\cite{yiping2017neurosurgeon,liu2020improve,shao2020communication}.
This is because the communication payload size of the DI is typically larger than that of local computing and cloud computing.
Moreover, the bandwidth of the IoT network is generally narrow; thus the communication latency is non-negligible in DI on IoT systems.

To achieve low communication latency, there are two general solutions: 1) adapting low-latency, which is a generally unreliable communication protocol,
 such as user datagram protocol (UDP) and higher physical transmission rate, and 2) lossy compression of the transmitted message.
To realize ultra-low-latency DI (e.g., lower than 10\,ms), it is necessary to simultaneously adopt both solutions.
However, there is a trade-off in the solutions between the communication latency and prediction accuracy of the DI.
The transport layer, for example, in the narrow-band and lossy IoT networks, the UDP transmission causes non-negligible packet losses, which degrades the inference accuracy by causing defects in the exchange of sub-DNN output between devices and edge servers.
In contrast, reliable communication protocol (i.e., transport control protocol (TCP) transmission) causes non-negligible communication latency due to the retransmissions of dropped packets. 
Moreover, the lossy compression reduces the redundancy of the message, which increases the negative effect of packet loss (i.e., degrades the accuracy) on the DI.

This motivated us to improve the trade-off between communication latency and prediction accuracy by efforts on ML techniques.
This study aims to design a DI method that achieves high accuracy using unreliable communication protocol on lossy IoT networks, where a considerable percentage of the transmitted packet is dropped.
To this end, we have proposed communication-oriented model tuning (COMtune) to achieve robustness against the packet loss due to the non-retransmission policy of the unreliable communication protocol.
Using COMtune, even when a part of the message exchanged between the nodes is dropped by the packet loss, one can obtain accurate inference results using the successfully received message.

To achieve such robustness against the packet loss, our key idea is to train the DNN through emulation of the effect of drops in the IoT network using the dropout technique~\cite{hinton2012improving}, which randomly drops the activation in DNN.
Through the training, the DNN would be able to provide accurate predictions using the dropped information.
Moreover, the dropout technique~\cite{hinton2012improving} is well known as a regularization method; 
thus, the DNN receives the benefits of the regularization effect, and simultaneously achieves robustness against packet loss.
Furthermore, to achieve even lower communication latency, COMtune employs lossy compression methods, which reduce the payload size of the message.
We should note that the lossy compression reduces the redundancy of the message, results in the degradation of the robustness to the packet loss; thus, the COMtune, which improves the robustness against the packet loss, has further significant role in achieving high accuracy when the compression is applied. 
The performance evaluation using the image classification task CIFAR-10 demonstrated that the COMtune achieved higher accuracy than existing methods under lossy communication links, even while employing lossy compression methods.

The contributions of this study are summarized as follows:
\begin{itemize}
	\item We have proposed COMtune to improve the trade-off in the DI framework, on the unreliable communication link between communication-latency and accuracy, using strong message compression.
	The message compression and robustness to the unreliable communication link are highly dependent on each other; the message compression can reduce the redundancy of the message, which further degrades the system robustness to the unreliable communication link.
	To the best of our knowledge, existing research has only addressed, either the message compression, or the robustness to the unreliable communication link.
	\item To improve the trade-off, COMtune tunes the DNN model by emulating the effects of the unreliable communication links using the dropout technique.
	The performance evaluation using CIFAR-10 demonstrated that the COMtune achieved higher accuracy than existing methods, under unreliable communication links even while employing lossy compression methods.
\end{itemize}

This study is an expanded version of~\cite{itahara2021packet} and evaluates the performance of COMtune when message compression is applied, and reveals that the COMtune is more efficient when the message compression is combined.

\begin{table*}[t!]
	\caption{
		Distributed inference frameworks toward low communication latency
	}
	\label{table:related_works}
	\centering
	\begin{tabular}{cccccccc}
		\toprule
		Name
		                                                                                                                                                                                          & Analog or digital & Communication reliability & Communication link Model & Approach \\
		\midrule
		\cite{shao2020bottlenet++,alite2021xie,krouka2021communication}                                                                                                                           & 
		Analog                                                                                                                                                                                    & -                 & -                         & -                                   \\
		\cite{yiping2017neurosurgeon,teerapittayanon2017distributed,shi2019improving,li2018jalad,matsubara2021neural,huang2020dynamic,koda2020communication,jankowski2020joint,matsubara2020head} & 
		Digital                                                                                                                                                                                   & Reliable          & -                         & -                                   \\
		\cite{shao2020bottlenet++}                                                                                                                                                                & 
		Digital                                                                                                                                                                                   & Unreliable        & one hop wireless          & ML training                         \\
		\cite{dhondea2021caltec,bragilevsky2020tensor,bajic2021latent}                                                                                                                            & 
		Digital                                                                                                                                                                                   & Unreliable        & End-to-end                & Tensor completion                   \\
		Proposed method                                                                                                                                                                           & 
		Digital                                                                                                                                                                                   & Unreliable        & End-to-end                & ML training                         \\
		\bottomrule
	\end{tabular}
\end{table*}

Correspondingly, however, independent of this work, a similar concept to improve the trade-off between unreliable communication and prediction accuracy through training of DNN by emulating the effect of unreliable communication has been presented in~\cite{shao2020bottlenet++}.
Meanwhile, there are two primary differences between \cite{shao2020bottlenet++} and our research, that is the
communication link assumption and model training scheme.
This study focuses on the end-to-end communication link and assumes packet loss, while \cite{shao2020bottlenet++} focuses on one-hop wireless links and assumes bit-error.
Thus, the proposed COMtune can be applied to any network that experiences packet loss due to queue or buffer overflow, as well as bit errors.
Second, our model training procedure is comparatively simpler to implement and more efficient in terms of accuracy.
This is because \cite{shao2020bottlenet++} uses custom non-differentiable functions in DNN to emulate the effect of the unreliable communication.
This procedure increases the implementation cost and decreases model training efficiency.
In contrast, our training methods only utilize a dropout layer for the emulation; 
thus, the proposed method is easier to implement and can accommodate the link emulation layer in the back-propagation process, 
which enables the model to benefit from the regularization effect caused by the model training using the dropout.

\section{Related works}
\label{sec:related_works}

This section summarizes the existing research that addresses the problems of communication overhead in DI.
The summary of the related works is given in Table~\ref{table:related_works}.
First, without specifying the DI framework, a vast majority of research~\cite{parvez2018low,Nasrallah2019ultra} has been addressed to improve the trade-off between the latency and reliability of the communication systems. 
The proposed COMtune is orthogonal to these researches and improves the trade-off beyond the limits of those improved by the efforts on the communication system~\cite{parvez2018low,Nasrallah2019ultra}.  

In DI frameworks, the inference task generated in the IoT device is offloaded to the other nodes by sharing the raw inputs, or the results of the local computation, which are referred to as vertical and horizontal DIs, respectively.
Unlike the research on vertical DI ~\cite{kim2017splitnet,zhao2018deeptings}, we have focused on the horizontal DI, because the sharing of the raw input in the vertical DI  includes a critical privacy risk.
In the horizontal DI literature, some works have addressed the achievement of low communication latency
by optimizing the division point~\cite{yiping2017neurosurgeon,shi2019improving,li2018jalad}, leveraging multiple sink nodes~\cite{teerapittayanon2017distributed}, pruning the DNN model~\cite{shi2019improving}, quantizing the message~\cite{li2018jalad,matsubara2021neural,huang2020dynamic}, dimensional reduction of the message~\cite{koda2020communication,jankowski2020joint,matsubara2020head}, and combining multiple inference tasks into a single one~\cite{matsubara2021neural}.
However, these works assumed a reliable communication link and aimed to reduce the communication payload size.
The problem of the trade-off between reliability and latency has not been addressed by these works; thus, they are orthogonal to this research.

Another direction is to adapt an analog communication system~\cite{shao2020bottlenet++,alite2021xie,krouka2021communication}.
\cite{alite2021xie,shao2020bottlenet++} used analog communication to reduce the cost of channel encoding in digital communication, where multiple nodes transmit signals in the same time slot by leveraging superimposition~\cite{krouka2021communication}.
However, this study focuses on digital communication, which is more widely used than analog communication. 

The impact of the unreliable communication link on DNN inference was evaluated in~\cite{liu2020improve}.
This work demonstrated the feasibility and effectiveness of employing unreliable but low-latency communication protocols for AI-empowered time-critical applications.
\cite{liu2020improve} transmits the raw input from the device to the server, which includes the critical privacy risks.
In the DI literature, to achieve robustness against the unreliable communication link, certain researches addressed estimation of the clean transmitted message from the received message, which is corrupted by the unreliable communication link, by joint source-channel coding \cite{choi2019neural}, linear tensor completion~\cite{dhondea2021caltec}, low-rank tensor completion~\cite{bragilevsky2020tensor}, and image inpainting based completion~\cite{bajic2021latent}.
Orthogonal to these works, which estimate the clean message from the corrupted message,
we aimed to achieve a split model that achieves highly accurate predictions from the corrupted message, and proposed a joint model training method.

\section{Proposed Method: Communication-oriented Model Fine-tuning}
\label{sec:proposed_method}
\begin{figure}[!t]
	\centering
	\includegraphics[width = 0.35\textwidth]{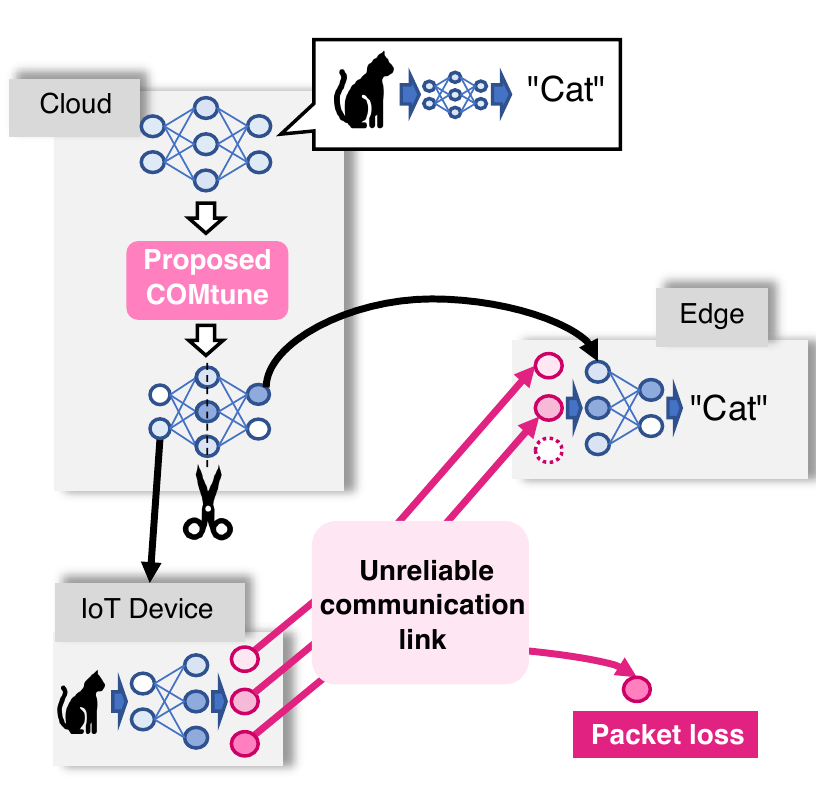}
	\caption{
		Over view of distributed inference with the proposed communication-oriented model tuning.
		The red arrows indicate the upload of the message from the device to the edge server via the unreliable communication link, in which the message is corrupted.
		The edge server obtains the prediction results using the corrupted message.
	}
	\label{fig:abst}
\end{figure}

\begin{figure}[!t]
	\centering
	\subfloat[Proposed: communication-oriented model tuning]{\includegraphics[width = 0.4\textwidth,page = 1]{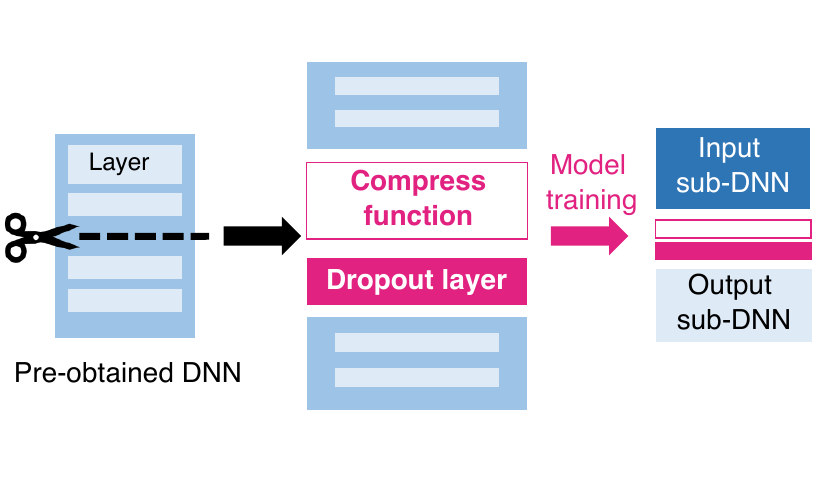}}\\
	\subfloat[Subsequently distributted inference]{\includegraphics[width = 0.4\textwidth,page = 2]{intro.pdf}}
	\caption{
		Detailed procedure of the proposed communication-oriented model tuning.
		Red arrows indicate the upload of the activation from the device to the edge server via the unreliable link, which does not retransmit dropped packets.
		The edge server obtains prediction results using only the successfully transmitted activations.
	}
	\label{fig:proposed_method}
\end{figure}

\subsection{System Model}
We assume an application scenario of automated surveillance in public places, roadsides, or factories, where IoT devices and edge servers cooperatively predict accidents, such as collisions, to avoid their occurrence.
IoT devices equipped with cameras, monitor the target area and send information to the edge server.
Based on the information, the edge server detects objects and their movement and further predicts the probability of the accident.
In the application scenario, latency is a critical issue because the edge server is required to complete the prediction before the incident occurs.

Fig.~\ref{fig:abst} shows the system model consisting of a cloud server, an edge server, and an IoT device.
The edge server and IoT devices are connected via lossy and narrow-band networks, and the messages are exchanged using an unreliable communication link.
The communication model is defined in more detail in the next section.
The cloud server obtains a pre-obtained DNN model from public repositories that are suitable for the inference tasks generated on the IoT device, for example, VGG~\cite{simonyan2014very} for image recognition tasks, YOLO~\cite{redmon2016you} for object detection tasks, and BERT~\cite{devlin2018bert} for neural language tasks.
As shown in Fig.~\ref{fig:proposed_method} (a),  the pre-obtained model is fine-tuned by the proposed COMtune method to provide accurate inference while conducting DI via the unreliable protocol under lossy and narrow-band networks with ultra-low latency.
The detailed COMtune procedure has been explained in Section~\ref{ssc:dropout}.
Following the fine-turning, the DNN is divided at a division layer and the portions (sub-DNNs) are distributed to the IoT device and the edge server.

As shown in Fig.~\ref{fig:proposed_method} (b), when an inference task is generated in the IoT device, the device and the server collaboratively solve the inference task using the distributed sub-DNNs, as follows:
The IoT device generates activation by feeding the input sample to the input sub-DNN, compresses the activation, and sends the compressed activation to the edge server via the unreliable communication link.
In the unreliable communication link, a non-negligible amount of packets are dropped; however, the dropped packets are not retransmitted.
The edge server obtains the prediction results through the output sub-DNN by inputting the successfully received activation from the IoT device.
The detailed DI procedure has been explained in Section~\ref{ssc:DI}.

\subsection{Unreliable Communication Link Assumption}
We assumed that the transmitted messages are probabilistically dropped owing to the non-retransmission policy of the unreliable communication protocol, where one does not retransmit the packets even when the packets are dropped. 
More formally, considering that the device sends a vector $\bm{x}$ via the communication link with a packet loss rate $p$,
the edge server successfully receives a vector $f^\mathrm{c}(\bm{x} \mid p)$ denoted as follows:
\begin{align}
	\label{equ:communication_channel}
	f^\mathrm{c}(\bm{x} \mid p) = \bm{x} \odot \bm{m}(p),
\end{align}
where operator $\odot$ indicates the element-wise product and $\bm{m}(p)$ is a binary vector following the Bernoulli distribution with an expected value of $1-p$.

In a real-world communication system, the vector of the activation $\bm{x}$ is divided into multiple packets and transmitted. 
Therefore, when a packet is dropped, the consecutive elements of $\bm{x}$ are lost. 
To avoid the burst loss, the device shuffles the vector elements and stores them in packets.
The edge server constructs vector $\bm{x}$ from the successfully received packets, which results in~\eqref{equ:communication_channel}.
That is, the device permutes the elements randomly and stores them in packets. 
A packet $\bm{p}_i$ is represented as follows:
\begin{align}
	\bm{p}_i & \coloneqq \{x_{k_j} \mid i \leq  j < i+s\},
\end{align}
where $k_j$ and $s$ are the permuted identification of the element and the number of elements stored in a packet, respectively. 
The edge server reconstructs the vector of activations from a subset of transmitted packets $\bm{P}^\mathrm{r}$, where
\begin{align}
	\bm{P}^\mathrm{r} = \{\bm{p}_i \mid \bm{p}_i\, \text{is received successfully}\}. 
\end{align}
Thus, the reconstructed vector is expressed as $\bm{x} \odot \bm{m}(p)$.

Assuming the aforementioned communications model, the number of the received packets and the latency are denoted as follows.
In the unreliable communication link, if $n^\mathrm{t}$ packets are transmitted using the communication link with a packet loss rate of $p$,
the probability mass function (PMF) of the number of received packets $n^\mathrm{r}$ is expressed as follows:
\begin{align}
	\mathrm{PMF}(n^\mathrm{r}) =
	\begin{cases}
		{n^\mathrm{t} \choose n^\mathrm{r}} p^{n^\mathrm{t}-n^\mathrm{r}}(1-p)^{n^\mathrm{r}}, \text{if}\ 0 \leq n^\mathrm{r} \leq n^\mathrm{t}; \\
		0,                                                                                      \text{otherwise}.
	\end{cases}
\end{align}
The expected number of received packets is denoted by $(1-p)n^\mathrm{t}$.
Assuming throughput $b$ and packet size $l$, the latency is calculated as $n^\mathrm{t}l/b$.
In contrast, all the transmitted packets are received when using a reliable communication link; thus, $n^\mathrm{r} = n^\mathrm{t}$.
The PMF of latency is
\begin{align}
	\label{equ:prev_delay}
	\mathrm{PMF}(\tau) =
	\begin{cases}
		{\lceil\tau/T\rceil  -1 \choose n^\mathrm{t}-1} p^{\lceil\tau/T\rceil-n^\mathrm{t}}(1-p)^{n^\mathrm{t}}, & \text{if}\ \lceil\tau/T\rceil \geq  n^\mathrm{t}; \\
		0,                                                                                                       & \text{otherwise}.
	\end{cases}
\end{align}

\subsection{Details of Communication-oriented Model Tuning}
\label{ssc:dropout}
To achieve high accuracy DI prediction under an unreliable communication link in highly lossy networks with activation compression,
the pre-obtained DNN is trained through emulation of the effect of packet loss and lossy activation compression.
The overview of the COMtune is depicted in Fig.~\ref{fig:proposed_method} (a).
To emulate the effect of packet loss, we determined that the behavior of the dropout layer is similar to the effect of the packet loss in the unreliable communication link, as defined in~\eqref{equ:communication_channel}, and used the dropout layer to emulate the effect of the packet loss.
Further, the dropout layer and the activation compression function are inserted into the division layer of the pre-obtained model. Subsequently, the model with the dropout layer and the activation compression is trained.

First, the cloud server obtains a pre-trained DNN model from the public repository, which is denoted by $f^\mathrm{pre}(\cdot\mid \bm{w}^\mathrm{pre})$, where $\bm{w}^\mathrm{pre}$ are the parameters.
The pre-trained DNN is divided into input-sub DNN $f^\mathrm{in}(\cdot\mid \bm{w}^\mathrm{in})$ and output-sub DNN $f^\mathrm{out}(\cdot\mid \bm{w}^\mathrm{out})$ as 
\begin{align}
	f^\mathrm{pre}(\cdot \mid \bm{w}^\mathrm{pre}) =
	f^\mathrm{out}(\cdot\mid \bm{w}^\mathrm{out}) \circ f^\mathrm{in}(\cdot \mid \bm{w}^\mathrm{in}),
\end{align}
where $f(\cdot) \circ g(\cdot)$ denotes the composite function of $f(\cdot)$ and $g(\cdot)$.
We tuned the DNN $f^\mathrm{trn}(\cdot \mid \bm{w}^\mathrm{trn})$, which consists of two sub-DNNs, the dropout layer, and compression functions.
The following section details the DNN $f^\mathrm{trn}(\cdot \mid \bm{w}^\mathrm{trn})$. 
Following the training, the input sub-DNN $f^\mathrm{in}(\cdot\mid \bm{w}^\mathrm{in})$ is sent to the IoT device, 
and an output sub-DNN $f^\mathrm{out}(\cdot\mid \bm{w}^\mathrm{out})$ is sent to the edge server, respectively.

The dropout was originally proposed as a regularization method in DNN literature, which enables training of the DNN for longer periods without overfitting and improves the test accuracy~\cite{hinton2012improving}. 
Thus, the dropout technique has been used in various DNN architectures and is available in multiple deep learning frameworks. 
In each training iteration using the dropout technique, the outputs of the hidden units are set to zero using a dropout layer with a dropout rate $r$. 
In addition to omitting the hidden unit outputs, the surviving (non omitted) hidden units are multiplied by $1/(1-r)$.
Hence, the dropout behavior $f^\mathrm{d}(\cdot \mid r)$ is represented as follows:
\begin{align}
	\label{equ:dropout}
	\bm{x}_{i+1} = f^\mathrm{d}(\bm{y}_i \mid r) =
	\frac{1}{1-r}\bm{y}_i \odot \bm{m}(r),
\end{align}
where $\bm{y}_i$ is the hidden unit of the $i$th layer, and $\bm{x}_{i+1}$ is the input of the $i+1$th layer.
Comparing equations~\eqref{equ:communication_channel} and~\eqref{equ:dropout},
we determine that the dropout technique can emulate the drops of activation due to packet loss, in the model training.
Therefore, the model trained using the dropout technique can provide accurate inferences even when the activations are dropped.

In addition to the unreliable communication link, the lossy compression reduces communication latency; however, it may degrade inference accuracy. 
To adapt the DNN model to the activation compression, COMtune fine-tunes the DNN model by inserting the compression function and dropout layer to the division layer.
In this study, we used either of the two general lossy compression methods, quantization and dimensional reduction, which are detailed in Appendix~\ref{ssec:compression}. Here, we have described COMtune with the general compression method.
The compression and decompression function are denoted by $f^\mathrm{cmp}(\cdot)$ and $f^\mathrm{dec} (\cdot)$, respectively.
Given the raw activation as $\bm{a}^\mathrm{raw}$, the compressed activation $\bm{a}^\mathrm{cmp}$ is denoted as $\bm{a}^\mathrm{cmp} \coloneqq f^\mathrm{cmp}(\bm{a}^\mathrm{raw} \mid M)$, where, $M$ is the data size of the compressed activation.
From the compressed activation, the uncompressed activation is estimated by $\bm{a}^\mathrm{dec} =  f^\mathrm{dec}(\bm{a}^{\mathrm{cmp'}}).$
Therefore, using the above defined functions, the DNN $f^\mathrm{trn}(\cdot \mid \bm{w}^\mathrm{trn})$ that fine-tuned in the COMtune is denoted as follows:
\begin{align}
	f^\mathrm{trn}(\cdot \mid \bm{w}^\mathrm{trn}) = f^\mathrm{out} & (\cdot\mid \bm{w}^\mathrm{out}) \circ f^\mathrm{dec}(\cdot) \notag                \\
	\circ f^\mathrm{d}(\cdot \mid r)                                      & \circ f^\mathrm{cmp}(\cdot \mid M)\circ f^\mathrm{in}(\bm{x} \mid \bm{w}^\mathrm{in}).
\end{align}
where $r$ is a dropout rate.

We should further note that the dropout rate and message size used in the model training corresponds to the packet loss rate and the message size in the DI procedure.
Thus, training using a larger dropout rate implies that the DNN is trained to adapt to a more lossy communication link, thus improving packet loss tolerance.
Training with a smaller message size in the fine-tuning implies adapting to use a smaller message size in the DI, thus reducing communication latency.
In contrast, as mentioned in~\cite{hinton2012improving}, a larger training dropout rate degrades the achievable model performance (i.e., performance without any packet loss); similarly, a smaller message size degrades the achievable model performance, as well \cite{courbariaux2016binarized}.
Therefore, the dropout rate and the message size are selected based on the packet loss rate of the communication link, desired communication latency, and model performance requirements.

Moreover, the dimensional reduction may strongly degrade the accuracy than quantization in highly unreliable communication link.
This is because, in the dimensional reduction, this paper adopts principal component analysis (PCA) to compress the message; the message is represented by a linear combination of the small number of basis vectors, and the coefficients of basis vectors are transmitted as the compressed message, leading to a significant difference in the contribution of each element of the compressed message, which is detailed in Appendix.
Thus, when the elements of the compressed message that correspond to important principal components (e.g., first principal components) are dropped, the accuracy is significantly degraded.
On the other hand, in the quantization, an element of the compressed message corresponds to an element of an uncompressed message.
Thus, the difference in the contribution between elements in the quantization is smaller than that in the dimensional reduction; this is a reason for the robustness of the quantization against packet loss.

\subsection{Details of Distributed Inference}
\label{ssc:DI}
The DI is conducted when an inference task with input $\bm{x}$ is generated in the IoT device, which is depicted in Fig.~\ref{fig:proposed_method} (b).
First, the device generates and compresses the activation as follows:
\begin{align}
	\bm{a} = f^\mathrm{cmp}(\cdot \mid M)\circ f^\mathrm{in}(\bm{x} \mid \bm{w}^\mathrm{in}).
\end{align}
Subsequently, the activation is transmitted by the communication link denoted in~\eqref{equ:communication_channel}.
Thus, the reconstructed vector is calculated as
\begin{align}
	\bm{a}' = f^\mathrm{c}(\bm{a} \mid p) = \bm{a} \odot \bm{m}(p).
\end{align}
To compensate the drops of the activation, the activation is multiplied by $1/{1-p}$.
From the compensated activation, the uncompressed activation is estimated as
\begin{align}
	\bm{a}^\mathrm{r} = f^\mathrm{dec}\left(\frac{1}{1-p}\bm{a}'\right).
\end{align}
Subsequently, $\bm{a}^\mathrm{r}$ is fed to the output-sub DNN, and we obtain the prediction result.
The prediction result $\bm{y}$ can be written as
\begin{align}
	\bm{y} = f^\mathrm{trn}(\cdot \mid \bm{w}^\mathrm{trn}) = f^\mathrm{out} & (\cdot\mid \bm{w}^\mathrm{out}) \circ f^\mathrm{dec}(\cdot) \notag                \\
	\circ f^\mathrm{c}(\cdot \mid p)                                          & \circ f^\mathrm{cmp}(\cdot \mid M)\circ f^\mathrm{in}(\bm{x} \mid \bm{w}^\mathrm{in}).
\end{align}

If $f^\mathrm{d} (\cdot \mid r)$ in the model training is close to $f^\mathrm{c}(\cdot \mid p)$ in the DI, the model is expected to accurately predict from the corrupted activation.
Comparing \eqref{equ:dropout} and \eqref{equ:communication_channel}, when the parameter $r$ is similar to the parameter $p$, 
$f^\mathrm{d} (\cdot \mid r)$ is similar to $f^\mathrm{c}(\cdot \mid p)$.
Thus, when $r$ is similar to $p$, the COMtune is expected to improve the prediction accuracy from the corrupted activation.
Moreover, our evaluation revealed that even when the difference between $r$ and $p$ is large (e.g., $(r,p)= (0.5,0.0)$), 
the COMtune achieved higher accuracy than the previous DI. 

\section{Evaluation}
\subsection{Setup}
\label{ssec:setup}
\begin{figure}[!t]
	\centering
	\includegraphics[width=0.4\textwidth]{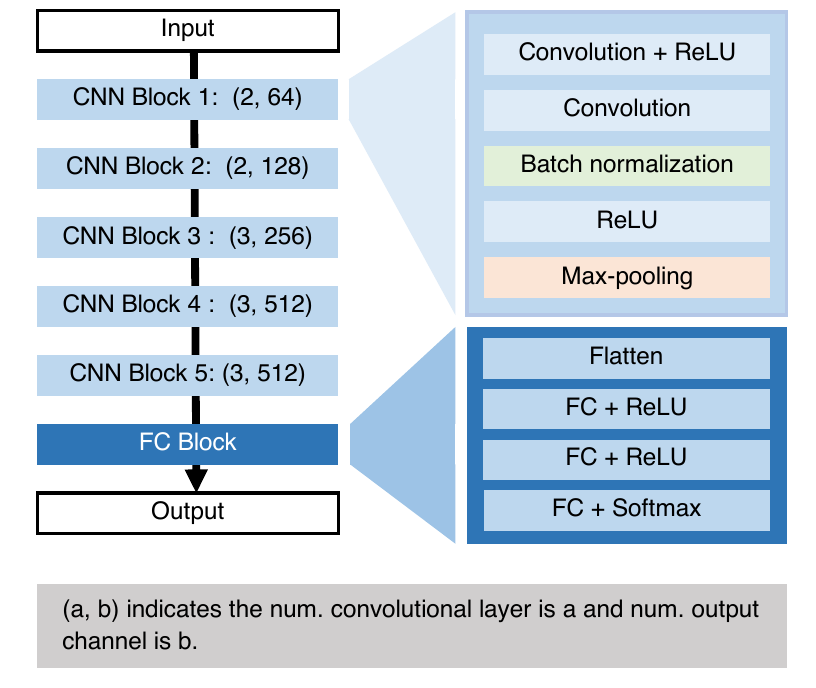}
	\caption{Architecture of DNN.
	Each convolutional neural network (CNN) block consists of two or three convolutional, batch normalization, and max-pooling layers.
	The number of convolutional layers $a$ in each CNN block and the output channel $b$ are denoted as $(a, b)$ in each CNN block.
	The fully connected (FC) block consists of three FC layers.}
	\label{fig:ML_model}
\end{figure}

\noindent\textbf{Communication setup.}
An IoT device and an edge server were assumed to be connected via a lossy IoT network, which was abstracted as a communication link, in which packets were randomly dropped with the probability $p$.
Hence, the elements of the activation vector transmitted by the IoT device were randomly dropped.
To calculate the communication latency, the packet size and throughput of the communication link (including MAC and network layer overheads) were set to 100\,bytes and 9.0\,Mbit/s.
We considered two communication protocols; unreliable protocol (i.e., without retransmissions) and reliable protocol (i.e., with retransmissions).

\noindent\textbf{Dataset and machine learning models.}
We used an image recognition dataset, CIFAR-10\footnote{\url{https://www.cs.toronto.edu/~kriz/cifar.html}}, with 50,000 training and 10,000 testing images that represented 10 image classes, such as ``dog'' and ``ship.''
The training dataset was used to fine-tune the pre-obtained model in the COMtune.
The test dataset was used to evaluate the inference performance of the DI phase.

The architecture of the DNN model used in the experiments is shown in Fig.~\ref{fig:ML_model}.
The model was designed with reference to VGG16~\cite{simonyan2014very}, which consists of five convolutional blocks and a FC block.
Each convolutional block included two or three $3 \times 3$ convolutional layers activated by the rectified linear unit (ReLU), and the block was followed by a $2 \times 2$ max-pooling layer.
The convolutional layers have the same number of output channels in each convolutional block.
Additionally, one of the two convolutional layers is followed by the batch normalization layer.
The FC block consists of three FC layers (256 and 128 units with ReLU activation and 10 other units activated by softmax).

\noindent\textbf{Machine learning training.}
The detailed ML training procedure is as follows: 
The training dataset is divided into updating and validation datasets in a ratio of 9:1.
The DNN model is updated using only the updating dataset for multiple epochs.
In each epoch, the model is evaluated using the validation dataset.
The training is completed if 150 epochs are performed, or if the validation loss increased after 20 epochs consecutively, which indicates that the model is starting to overfit.
The Adam optimizer, a training rate of 0.001, and a mini-batch size of 128 were selected as hyperparameters.
Notably, this paper generates a pre-obtained model by training a randomly initialed ML model using the aforementioned training procedure.  

\noindent\textbf{Distributed inference.}
In this evaluation, major parts of the inference task were offloaded to the edge server
 because the computational capacity of the IoT device is generally much worse than the edge server.
Specifically, the CNN was divided into CNN block 1, resulting in the inference tasks of CNN block 1 being conducted at the IoT device and that of the CNN blocks 2, 3, 4, and 5, and the FC block being conducted in the edge server.
The dimensions of the activation of the CNN block 1 is 16,384, which is 65.5\,kB in 32bit float point representation 
(e.g., the communication delay is 58.2\,ms when no packet loss occurs).
The packet loss in the communication link is emulated by the dropout, where the dropout rate is set to the packet loss rate, which ranges from 0 to 0.9.
Additionally, we ran each method ten times from different random seeds and computed the average and standard deviation of the performance in ten trials.

\subsection{Cumulative distribution function of the accuracy and latency}
\begin{figure}[!t]
	\centering
	\subfloat[Communication latency]{\includegraphics[width=0.45\textwidth]{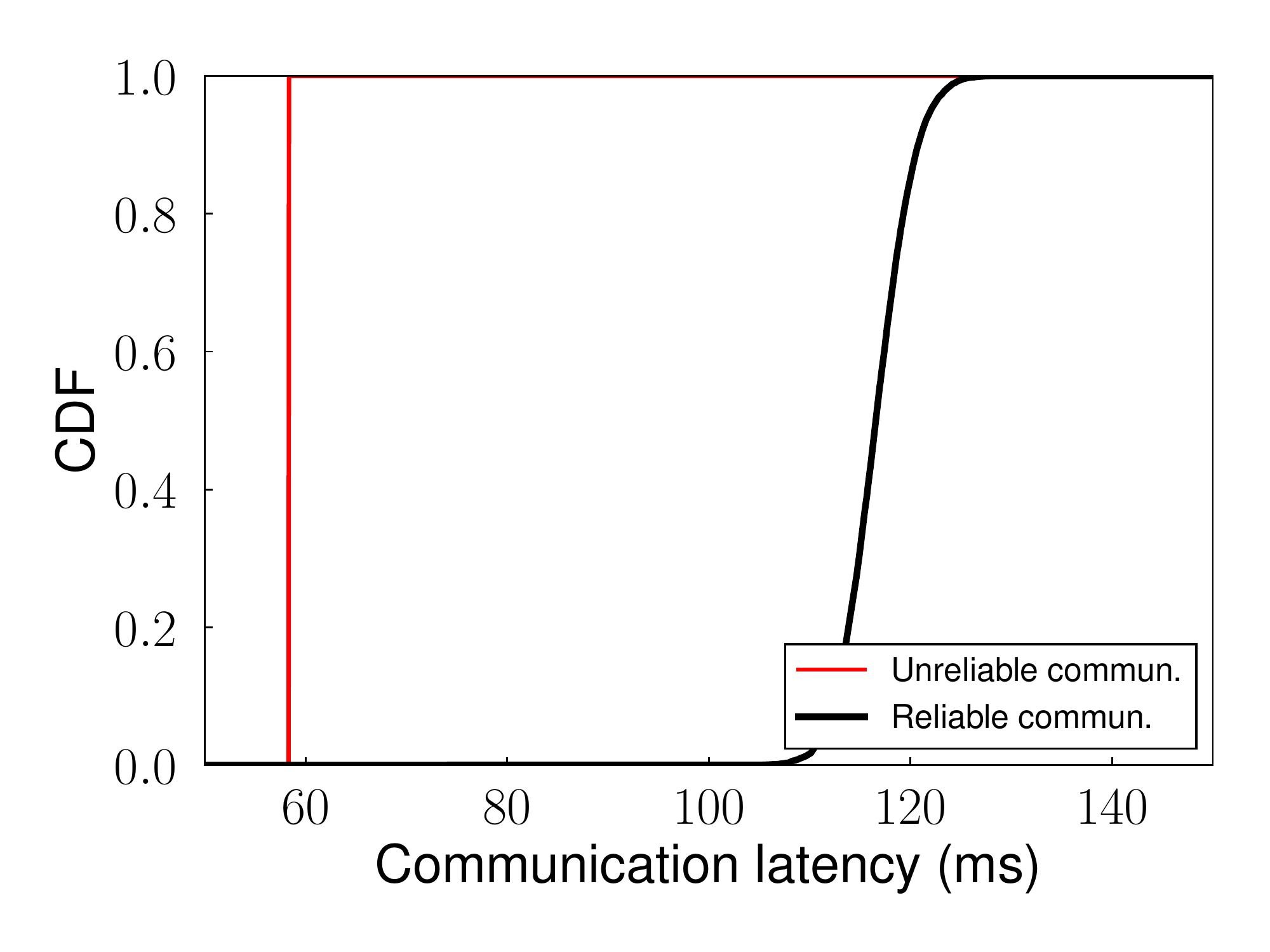}}\\
	\subfloat[Accuracy]{\includegraphics[width=0.45\textwidth]{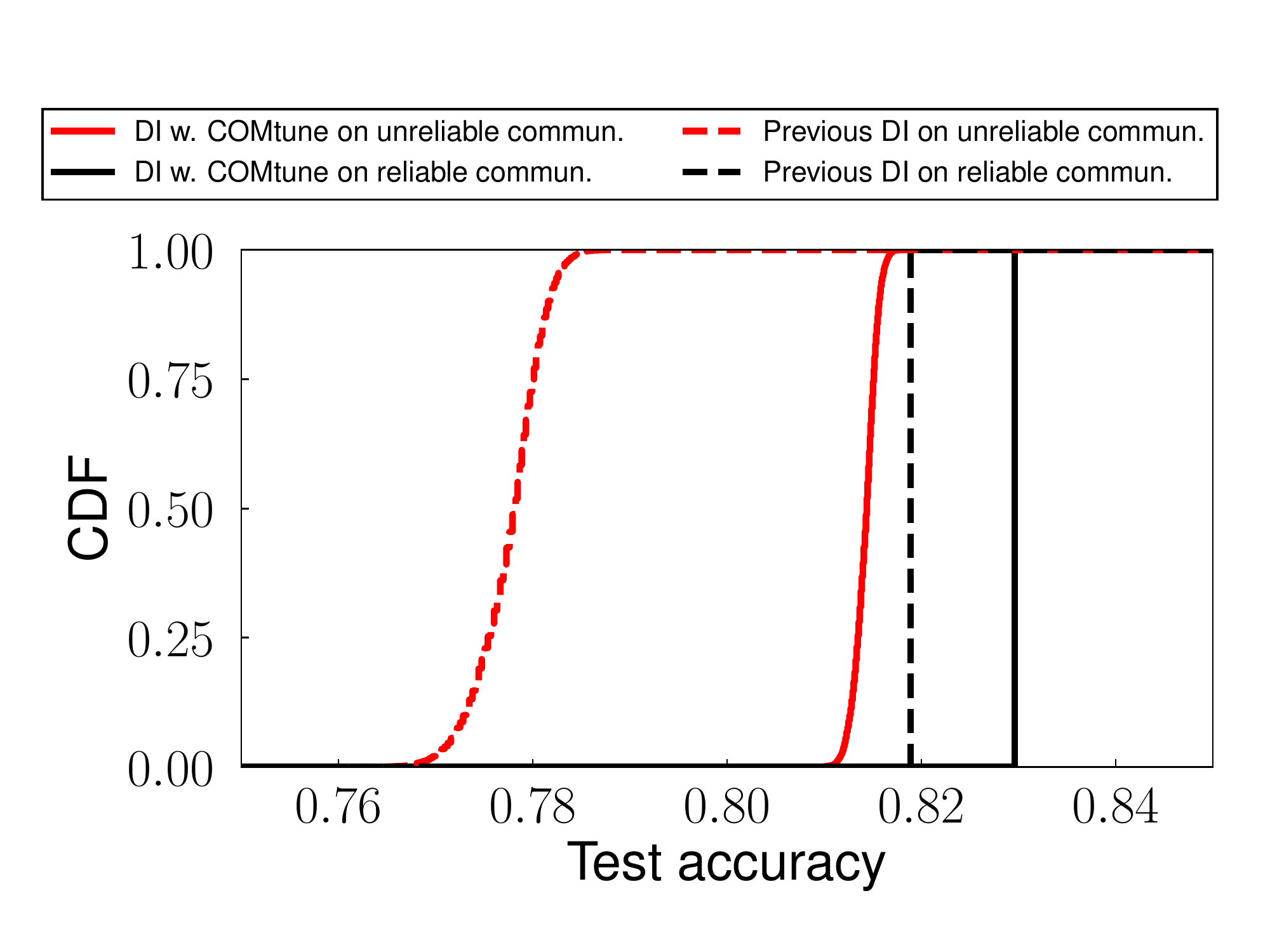}}
	\caption{Cumulative distribution function of the accuracy and communication latency for the previous DI and proposed DI with COMtune on the reliable and unreliable communication links, respectively.}
	\label{fig:prob_same_rx}
\end{figure}

Fig.~\ref{fig:prob_same_rx}~(a) illustrates the cumulative distribution function (CDF) of the communication latency of the DI using reliable and unreliable protocols, respectively, where the activation compression is not applied.
The CDF is obtained following the aforementioned discussion, with the parameters described in Section~\ref{ssec:setup} and the packet loss rate of 0.5. 
While using the unreliable protocol, 50\% of a message is dropped.
However, in case that a reliable protocol is used, the entire message is successfully received by retransmissions.
As shown in Fig.~\ref{fig:prob_same_rx}~(a), due to the no-retransmission policy, the latency of the unreliable protocol is stable and lower than that of the reliable protocol.
Moreover, the latency of the reliable protocol transmission is not stable.

Fig.~\ref{fig:prob_same_rx}~(b) shows the CDF of the accuracy of the proposed DI with COMtune, and previous DI using unreliable and reliable protocols, respectively.
Regardless of the underlying communication system, the proposed DI with COMtune achieved higher accuracy than the previous DI.
This is because of two reasons: regularization and robustness to the packet loss.
In the case of the reliable protocol, the accuracy of the DI with COMtune and previous DI is stable because all the transmitted packets are successfully received because of the retransmissions.
The DI with COMtune achieved 1\% higher accuracy than the previous DI because of the regularization effect of the dropout technique used in the COMtune.
In contrast, for the unreliable protocol, the accuracy of the DI with COMtune and previous DI is not stable due to the transmitted packets being dropped because of the non-retransmission policy of the unreliable protocol.
In the unreliable protocol transmission, the DI with COMtune achieved 4\% higher accuracy than the previous DI.
Moreover, comparing the accuracy degradation from that on the reliable protocol to that of the unreliable protocol,
the degradation of the DI with COMtune is smaller than that of the previous DI.
Thus, we can conclude that the COMtune improved the trade-off between the prediction accuracy and the communication latency,
due to the training involving emulation of corruptions of the message in the unreliable and low-latency communication system.  

\subsection{Impact of Dropout Rate on Robustness Against Packet Loss}
\label{ssec:main_result}
\begin{figure}[!t]
	\centering
	\includegraphics[width = 0.4\textwidth]{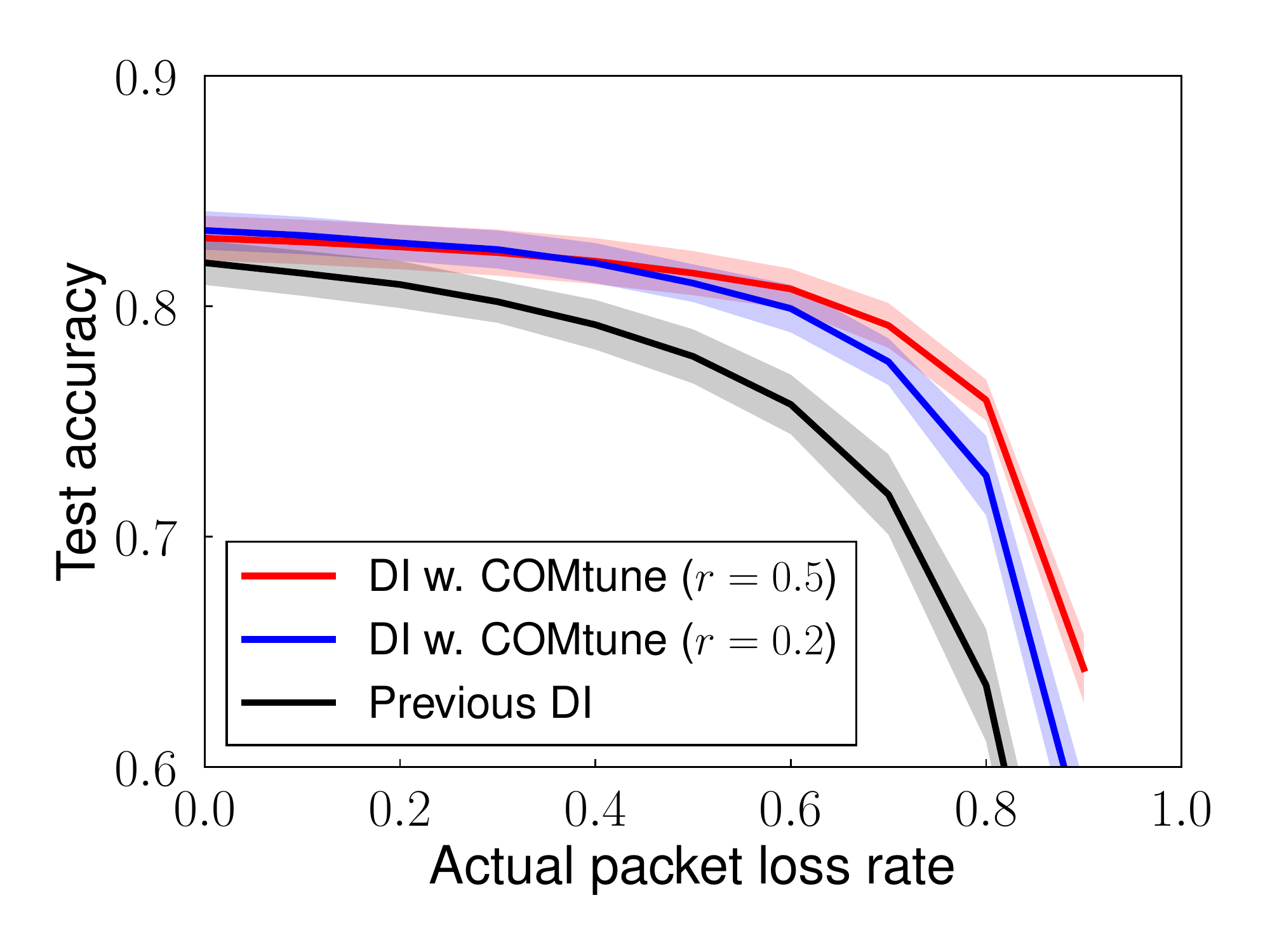}\\
	\caption{Test accuracy as a function of packet loss rate for each dropout rate $r$.
		The shaded regions denote the standard deviation of the performance among ten trials.
	}
	\label{fig:0704}
\end{figure}

Fig.~\ref{fig:0704} shows the test accuracy of the DI with COMtune, and previous DI as a function of the packet loss rate while using the unreliable communication link.
In the case of DI with COMtune, Fig.~\ref{fig:0704} shows the result for each dropout rate (i.e., 0.2 and 0.5) used in the COMtune prior to the DI. 
For both of the dropout rates, the DI with COMtune achieved higher accuracy than the previous DI regardless of the packet loss rate,
which indicates that COMtune improves the robustness of the split model against the packet loss.
Moreover, COMtune mitigates the accuracy degradation without any packet loss, as compared to the previous DI.
In particular, the accuracy of the previous DI was degraded by more than 10\%, when more than 70\% of the packets were dropped, 
while that of DI with COMtune with the dropout rate of 0.5 exhibited only a 3.8\% degradation in accuracy.
Thus, we can conclude that COMtune improves the packet loss robustness, even when the dropout rate in COMtune differs from the packet loss rate.

Moreover, as the dropout rate increases, the accuracy degradation is better mitigated.
Particularly, when the packet loss rate is 0.7, the model trained with a dropout rate of 0.5 and 0.2 demonstrated a 3.8\% and 5.7\% degradation in accuracy, respectively. 
This is because a larger dropout rate indicates emulation of the more lossy network in model training, which encourages the model to achieve high accuracy in the highly lossy network.

\subsection{Performance Evaluation with COMtune using Activation Compression}
\subsubsection{Effect of Activation Compression on Achievable Accuracy}
\begin{figure}[!t]
	\centering
	\includegraphics[width=0.4\textwidth]{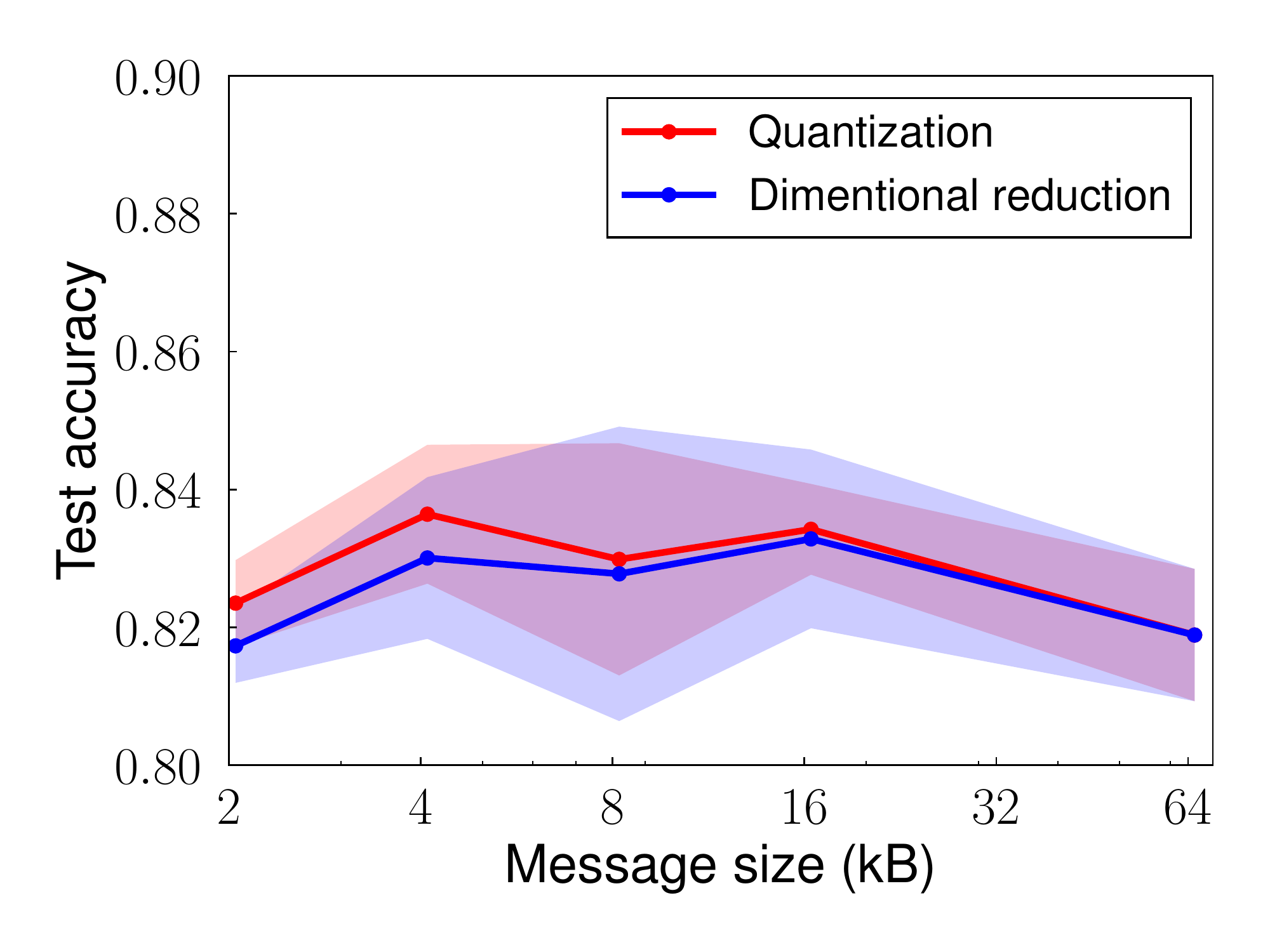}
	\caption{Effect of message compression on achievable accuracy.
		The message size without any compression is 64\,kB.}
	\label{fig:compress_fine}
\end{figure}

Fig.~\ref{fig:compress_fine} shows the test accuracy as a function of the message size without any packet loss, that is, that all transmitted packets were successfully received.
The message is compressed by either quantization or dimensional reduction, which are both detailed in Appendix~\ref{ssec:compression}.
Even when the message is compressed, the accuracy is comparable to that when the message is not compressed (i.e., the 65.5\,kB message),
which is consistent with the existing works that have addressed DNN compression~\cite{courbariaux2016binarized}.
However, the following evaluation, as shown in Fig.~\ref{fig:dr_comp}, reveals that there is a trade-off between the message size and robustness to the unreliable communication link; when the message is highly compressed, the robustness is degraded.  

\subsubsection{Effect of Communication-oriented Model Tuning on Accuracy with Activation Compression}
\begin{figure}[!t]
	\centering
	\subfloat[Quantization]{\includegraphics[width = 0.4\textwidth]{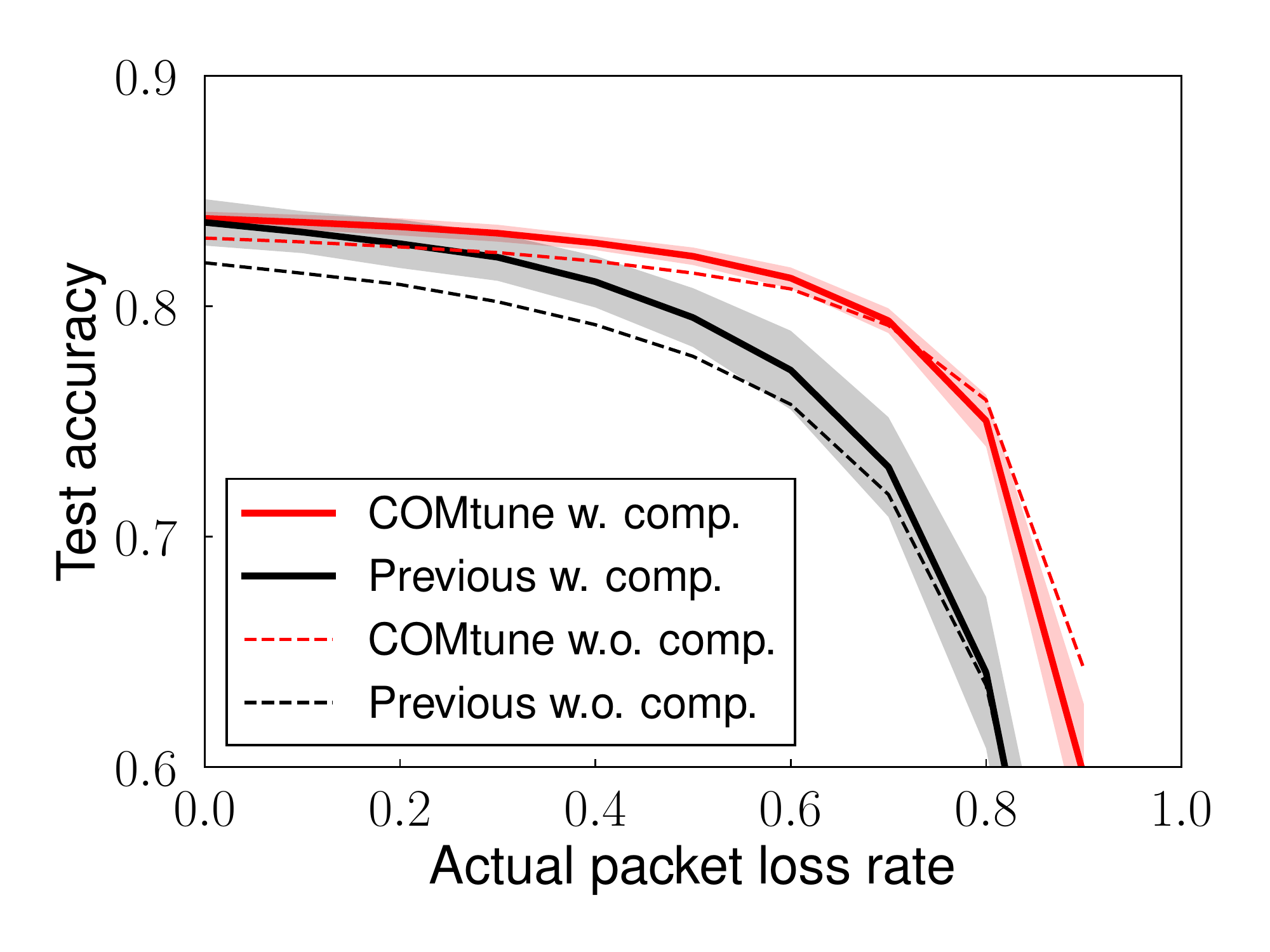}}\\
	\subfloat[Dimensional reduction]{\includegraphics[width = 0.4\textwidth]{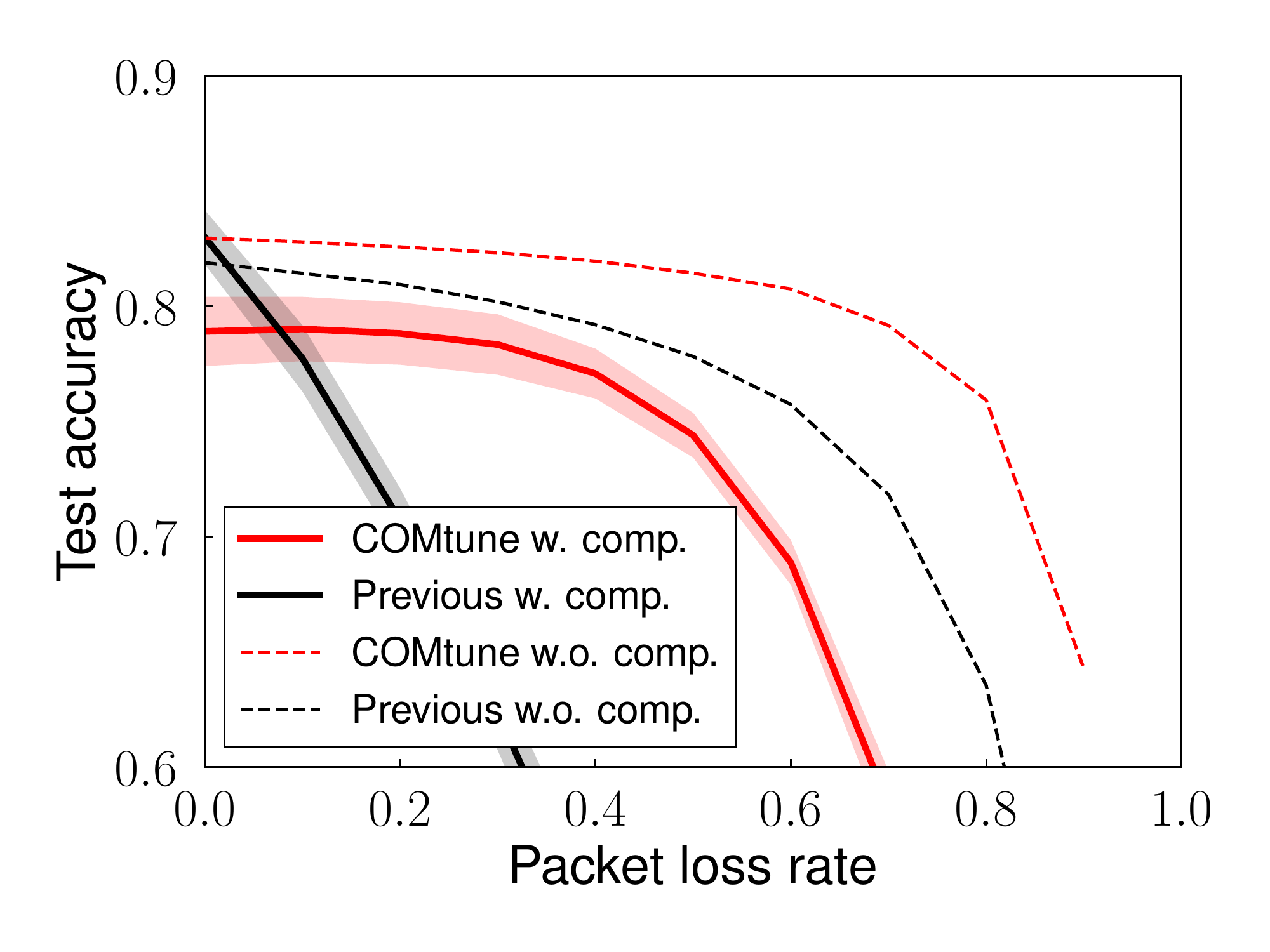}}
	\caption{Test accuracy as a function of packet loss rate with or without message compression (message size is 4\,kB and 64\,kB, respectively).
		The black and red lines indicate the results obtained using the DNN tuned without any dropout layer and the COMtune with dropout rates of 0.5, respectively.
		The solid lines and shaded regions denote the average and standard deviation of the accuracy among ten trials with the message compression, respectively.
		The dots lines indicate the average accuracy without message compression.
	}
	\label{fig:0711}
\end{figure}

Fig.~\ref{fig:0711} shows the test accuracy as a function of packet loss rate, with or without message compression (message size is 4\,kB and 64\,kB, respectively), using the unreliable protocol.
Fig.~\ref{fig:0711} (a) and (b) show the results when the quantization and dimensional reduction are applied to compress the message, respectively.
In Fig.~\ref{fig:0711} (a), when the compression is applied,
the accuracy of DI with COMtune is higher than that of the previous DI regardless of the packet loss rate, which is consistent with Fig.~\ref{fig:0704}.
This demonstrated that COMtune improved the packet loss tolerance of the split model in case that the message is highly compressed, as well as the message is not compressed. 
In Fig.~\ref{fig:0711} (b), the DI with COMtune achieved higher accuracy than precious DI when the dropout rate and the packet loss rates are similar.
In particular, DI with COMtune with the dropout rate of 0.5 does when the packet loss rate is larger than 0.1.

Comparing the accuracy with and without compression,
when the dimensional reduction is applied, the accuracy with compression is more degraded than without compression.
For example, the accuracy degradation when the packet loss rate of 0.5 is 7.0\% for DI with COMtune,  and that is 34.6\% for previous DI.
On the other hand, when the quantization is applied, the accuracy with compression is comparable to that with compression.
As discussed in Section~\ref{ssc:dropout}, this gap between the two message compression methods is explained in terms of the difference in the contribution of each element of the compressed message; the difference in dimensional reduction is significantly larger than the quantization. 
Thus, in the dimensional reduction, the accuracy is significantly degraded when the element of the compressed message has a high contribution.
Therefore, we can conclude that quantization is a message compression method that achieves more robustness against the packet loss than the dimensional reduction.

\begin{figure}[!t]
	\centering
	\includegraphics[width=0.4\textwidth]{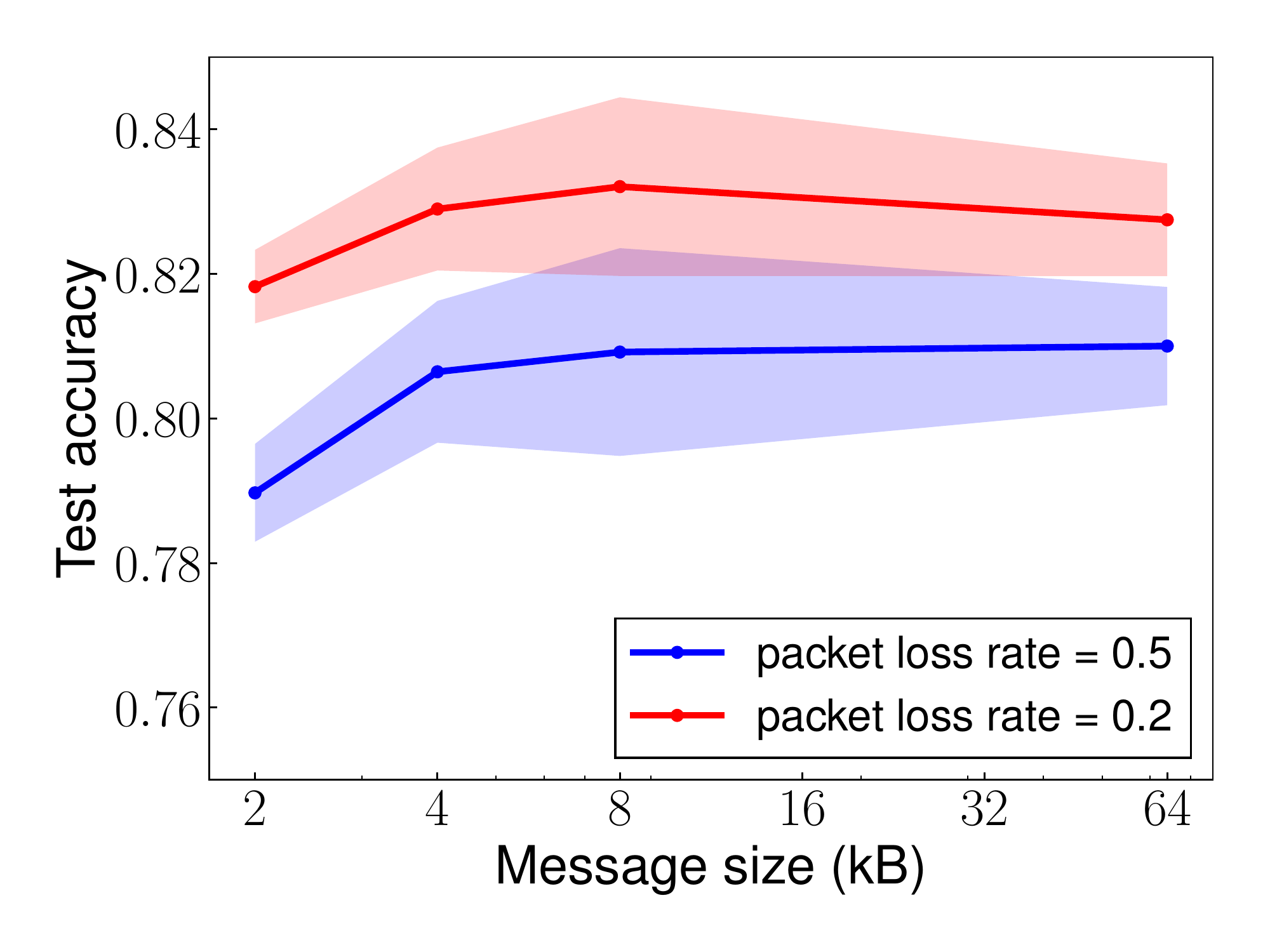}
	\caption{Effect of message size of DI with COMtune on robustness to the unreliable communication link.
		Quantization is applied as the message compression method, and DNNs are turned with dropout rates of 0.2.
	}
	\label{fig:dr_comp}
\end{figure}

Fig.~\ref{fig:dr_comp} shows the test accuracy of DI using COMtune as a function of the message size for the packet loss rate of 0.2 and 0.5, respectively.
In this evaluation, the quantization is applied as the message compression method.
Regardless of the packet loss rate, the accuracy is degraded as the message size is reduced.
Thus, we conclude that message compression degrades the robustness of the DI system to the unreliable communication link.
This is because message compression reduces the redundancy of the message.

\section{Conclusion}
\label{sec:conclusion}
We have presented COMtune that aims to improve the prediction accuracy and communication latency by efforts on the application layer.
Specifically, we aimed to achieve accuracy prediction under low-latency and unreliable communication link, such as UDP transmission.
In COMtune, the key idea is to train the ML model by emulating the effect of the unreliable communication link, such that the model gains robustness to the unreliable communication system.
Our experimental ML evaluation revealed that DI with COMtune obtains a more accurate prediction than previous DI on the highly unreliable communication link.
Moreover, we revealed that the proposed COMtune is compatible with the general message compression methods. 
An interesting area for future work is an optimization framework that determines the parameters of the emulated communication systems to maximize the model accuracy in lossy wireless networks under the constraints of the total latency of communication and computation.

\appendices

\bibliographystyle{IEEEtran}
\bibliography{main.bbl}

\section{Compression methods}
\label{ssec:compression}
To reduce communication payload size of the message (i.e., activation of the input-sub DNN), we adapted general lossy compression methods that are quantization and dimensional reduction.
Generally, the lossy compression method in DNN literature implies the compression of both, the activation and model parameters aiming to reduce the data size of the parameters and the computation cost of the inference.
However, this study aims to reduce the data size of the activation, thus only the activation is compressed.

\textbf{Activation quantization.}
In the quantization, the elements of the activation are compressed from full-perception values (i.e., 32 bit float representation) to quantized values (i.e., $n$ bit integer representation). 
The quantized activation is transmitted to the edge server as a message and dequantized to full-perception activation in the edge server, which is fed to output-sub network, in which the communication payload is reduced by $n/32$.
Given the desired message size $M$ and uncompressed message size $M^\mathrm{float}$, which is the message size with 32bit float point, $n$ is determined as $n = \lfloor 32M/M^\mathrm{float} \rfloor$. 

For more details of quantization, the elements are clipped into predefined ranges, and further represented by $n$ bit integers.
First, the full-perception activation elements are clipped into range from $s^\mathrm{min}$ to $s^\mathrm{max}$, where $s^\mathrm{min}$ and $s^\mathrm{max}$ are scale factors that indicate smallest and largest value represented by quantized value, respectively.
The scale factors are determined for each activation element based on the range of the distribution of the element using the pre-obtained dataset.
Finally, the clipped value is quantized to $n$ bit integer.

Hence, given $i$th element of full-perception activation as $a^\mathrm{float}_i$, the corresponding clipped value $a^\mathrm{clip}_i$ is denoted as
\begin{align}
	a^\mathrm{clip}_i =  \max \left(\min \left(a^\mathrm{float}_i, s^\mathrm{min}_i \right),s^\mathrm{max}_i \right).
\end{align}
Note that the scale factors are determined in the cloud server prior to the DI.
The quantized activation $a^\mathrm{int}_i$ is represented as
\begin{align}
	a^\mathrm{int}_i = \mathrm{round}\left(\frac{2^n -1}{s^\mathrm{max}_i-s^\mathrm{min}_i} a^\mathrm{float}_i\right).
\end{align}
For shorthand notation, we denote a quantization function of a single element of the activation by $f^\mathrm{qut}(a,s^\mathrm{min},s^\mathrm{max})$, where $f^\mathrm{qut}(a^\mathrm{float}_i,s^\mathrm{min}_i,s^\mathrm{max}_i) = a^\mathrm{int}_i$.
From the quantized activation $a^\mathrm{int}_i$, the unquantized activation is estimated as follows:
\begin{align}
	a^\mathrm{deq}_i = \frac{s^\mathrm{max}_i-s^\mathrm{min}_i}{2^n -1} a^\mathrm{int}_i.
\end{align}
For shorthand notation, we denote a dequantization function of a single element of the activation by $f^\mathrm{deq}(a,s^\mathrm{min},s^\mathrm{max})$, where $f^\mathrm{deq}(a^\mathrm{int}_i,s^\mathrm{min}_i,s^\mathrm{max}_i) = a^\mathrm{deq}_i$.
Thus, given $D$ dimensional vectors $\bm{s}^\mathrm{min}$, $\bm{s}^\mathrm{max}$, and $\bm{a}$ as the scale factors and uncompressed activation,
the compression and decompression functions are denoted as
\begin{align}
	f^\mathrm{cmp}(\bm{a} \mid M) & = \left\{f^\mathrm{qut}(a_i,s^\mathrm{min}_i,s^\mathrm{max}_i)\mid 0<i\leq D \right\}, \\
	f^\mathrm{dec}(\bm{a})       & = \left\{f^\mathrm{deq}(a_i,s^\mathrm{min}_i,s^\mathrm{max}_i)\mid 0<i\leq D \right\}.
\end{align}

\textbf{Dimensional reduction.}
In dimensional reduction, the activation is converted to a linear combination of basis vectors, where the number of the basis vectors is smaller than the dimensions of the activation.
In the DI, the coefficients of basis vectors are transmitted rather than the elements of the activation, which reduces the communication payload size by $D'/D$,
where the number of the basis vectors is $D'$ and the dimensions of the activation is $D$.
Thus, given the compressed message size $M$ and uncompressed message size $M'$,
$D'$ is determined as $D' =  \lfloor MD/M' \rfloor$. 
The server estimates the uncompressed activation using the basis vector and the received coefficients.
Formally, given a $D$ dimensional vector $\bm{a}$ as the uncompressed activation and $D'$ dimensional vector $\bm{a}'$ as a compressed activation,
the compression and decompression functions are denoted as 
\begin{align}
	f^\mathrm{cmp}(\bm{a}\mid M) & =   \bm{w}\bm{a},                      \\
	f^\mathrm{dec}(\bm{a}')       & =   \bm{w}^\mathrm{T}\bm{a}' + \bm{b},
\end{align}
where $\bm{w}$ is a $D'\times D$ matrix and $\bm{b}$ indicates $D$ dimensional bias vector, respectively.

To determine the parameters $\bm{w}$, PCA is used.
In more detail, $i$th row of $\bm{w}$ is an eigenvector of the data covariance matrix $\bm{S}$ of the pre-obtained dataset, which corresponds to $i$th largest eigenvalue.
The data covariance $\bm{S}$ is denoted as 
\begin{align}
	\bm{S}            & = \frac{1}{|\mathcal{A}|}\sum_{\bm{a}\in\mathcal{A}}(\bm{a}-\overline{\bm{a}})(\bm{a}-\overline{\bm{a}})^\mathrm{T}, \\
	\overline{\bm{a}} & \coloneqq \frac{1}{|\mathcal{A}|}\sum_{\bm{a}}\bm{a},
\end{align}
where
\begin{align}
	\mathcal{A} & = \{f^\mathrm{in}(\bm{x}_j \mid  \bm{w}^\mathrm{in}) \mid \bm{x}_j \in \text{preobtained dataset}\}.
\end{align}
The bias vector $\bm{b}$ is denoted as
\begin{align}
	\bm{b} = \sum_{i=D'+1}^{D} ({\overline{\bm{a}}^\mathrm{T}\bm{u}_i})\bm{u}_i,
\end{align}
where $\bm{u}_i$ is an eigenvector of $\bm{S}$, corresponding to $i$th largest eigenvalue.

\begin{IEEEbiography}
	[{\includegraphics[width=1in, height=1.25in, clip, keepaspectratio]{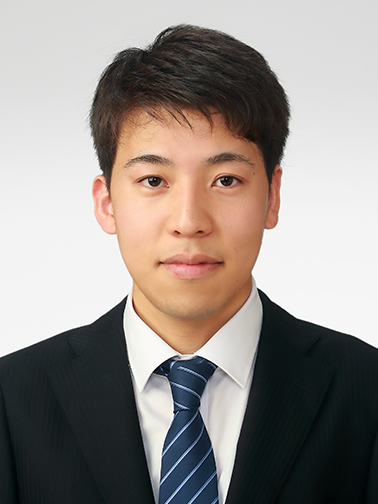}}]
	{Sohei~Itahara}
	received the B.E. degree in electrical and electronic engineering from Kyoto University in 2020.
	He is currently studying toward the M.I. degree at the Graduate School of Informatics, Kyoto University.
	He is a student member of the IEEE.
\end{IEEEbiography}

\begin{IEEEbiography}
	[{\includegraphics[width=1in, height=1.25in, clip, keepaspectratio]{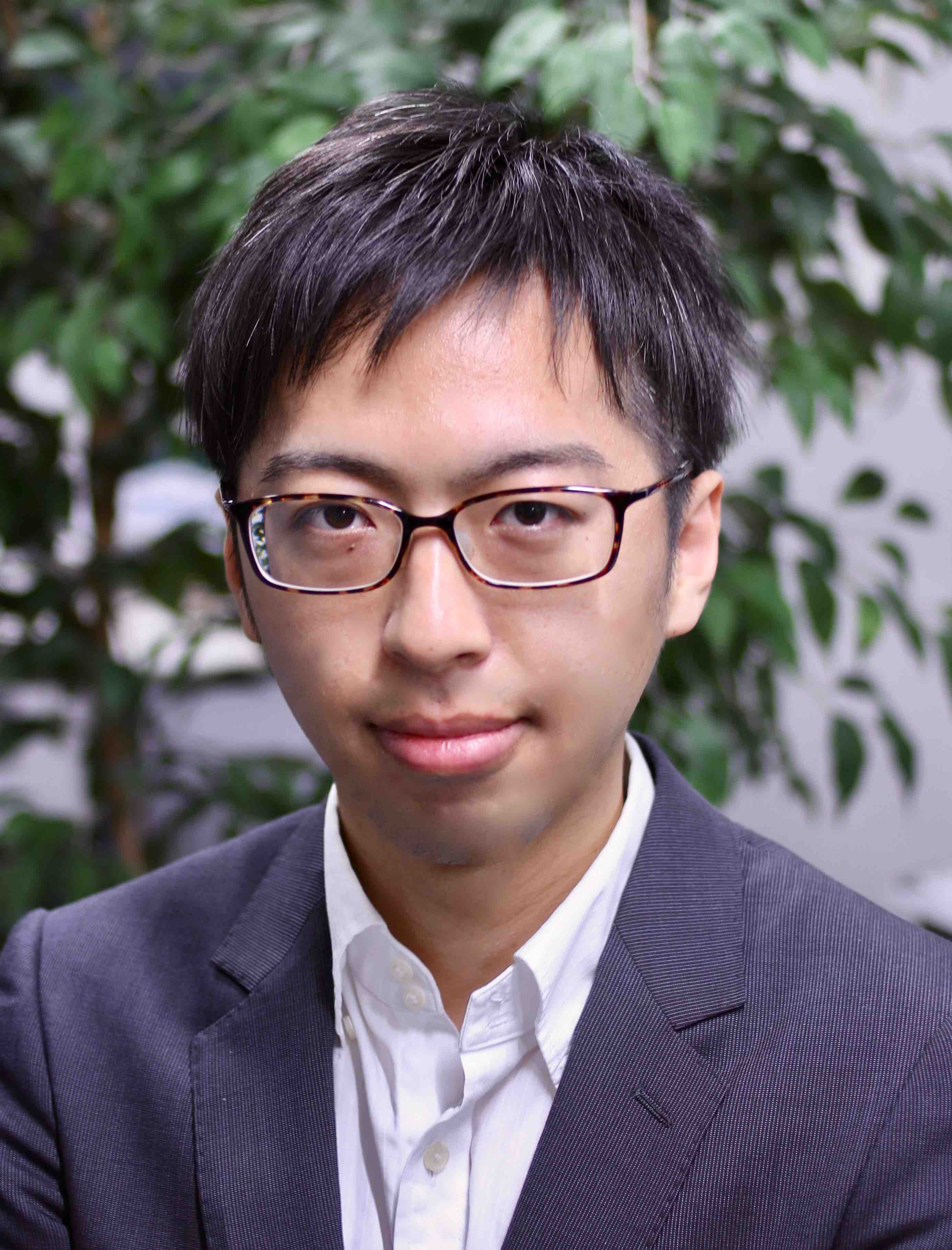}}]
	{Takayuki~Nishio}
	has been an associate professor in the School of Engineering, Tokyo Institute of Technology, Japan, since 2020.
	He received the B.E.\ degree in electrical and electronic engineering and the master's and Ph.D.\ degrees in informatics from Kyoto University in 2010, 2012, and 2013, respectively.
	He had been an assistant professor in the Graduate School of Informatics, Kyoto University from 2013 to 2020.
	From 2016 to 2017, he was a visiting researcher in Wireless Information Network Laboratory (WINLAB), Rutgers University, United States.
	His current research interests include machine learning-based network control, machine learning in wireless networks, and heterogeneous resource management.
\end{IEEEbiography}

\begin{IEEEbiography}[{\includegraphics[width=1in, height=1.25in, clip, keepaspectratio]{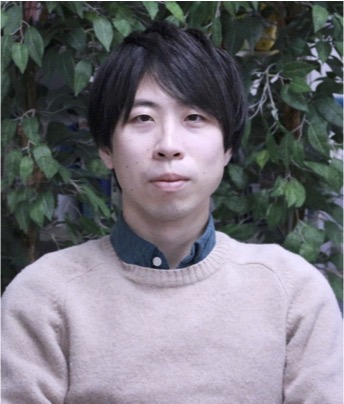}}]{Yusuke~Koda}
	received the B.E. degree in electrical and electronic engineering from Kyoto University in 2016 and the M.E. degree at the Graduate School of Informatics from Kyoto University in 2018.
	In 2019, he visited Centre for Wireless Communications, University of Oulu, Finland to conduct collaborative research.
	He is currently studying toward the Ph.D. degree at the Graduate School of Informatics from Kyoto University.
	He was a Recipient of the Nokia Foundation Centennial Scholarship in 2019.
	He received the VTS Japan Young Researcher's Encouragement Award in 2017.
	He is a member of the IEICE and a member of the IEEE.
\end{IEEEbiography}

\begin{IEEEbiography}
	[{\includegraphics[width=1in, height=1.25in, clip, keepaspectratio]{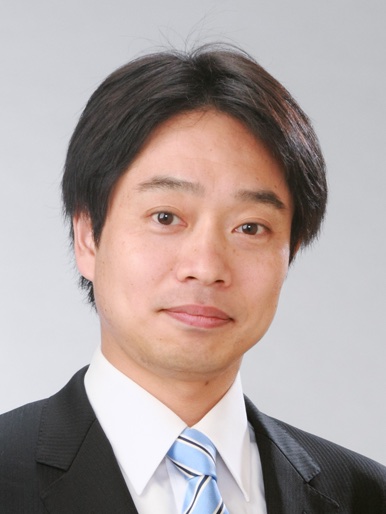}}]
	{Koji~Yamamoto}
	received the B.E. degree in electrical and electronic engineering from Kyoto University in 2002, and the M.E. and Ph.D. degrees in Informatics from Kyoto University in 2004 and 2005, respectively.
	From 2004 to 2005, he was a research fellow of the Japan Society for the Promotion of Science (JSPS).
	Since 2005, he has been with the Graduate School of Informatics, Kyoto University, where he is currently an associate professor.
	From 2008 to 2009, he was a visiting researcher at Wireless@KTH, Royal Institute of Technology (KTH) in Sweden.
	He serves as an editor of IEEE Wireless Communications Letters from 2017 and the Track Co-Chairs of APCC 2017 and CCNC 2018.
	His research interests include radio resource management and applications of game theory.
	He received the PIMRC 2004 Best Student Paper Award in 2004, the Ericsson Young Scientist Award in 2006.
	He also received the Young Researcher's Award, the Paper Award, SUEMATSU-Yasuharu Award from the IEICE of Japan in 2008, 2011, and 2016, respectively, and IEEE Kansai Section GOLD Award in 2012.
\end{IEEEbiography}

\EOD
\end{document}